\documentclass[preprint,12pt]{elsarticle}

 \usepackage{url} 
 \usepackage{algorithm}
\usepackage[noend]{algpseudocode} 
\usepackage{amsmath}
\usepackage{amssymb}
\usepackage{amsmath}
\usepackage{booktabs}
\usepackage{xcolor}

\journal{Nuclear Physics B}

\begin{document}

\begin{frontmatter}



\title{Fact or Fake? Assessing the Role of Deepfake Detectors in Multimodal Misinformation Detection}


\author[label1]{A S M Sharifuzzaman Sagar}
\author[label1]{Mohammed Bennamoun}
\author[label1]{Farid Boussaid}
\author[label1]{Naeha Sharif}
\author[label1]{Lian Xu}
\author[label2]{Shaaban Sahmoud}
\author[label3]{Ali Kishk}
\affiliation[label1]{organization={The University of Western Australia},
            city={Perth},
            country={Australia}}

\affiliation[label2]{organization={Fatih Sultan Mehmet Vakif University},
            city={Istanbul},
            country={Turkey}}

\affiliation[label3]{organization={Aljazeera Media Network Investigative Department},
            state={Doha},
            country={Qatar}}


\begin{abstract}

In multimodal misinformation, deception usually arises not just from pixel-level manipulations in an image, but from the \textit{semantic and contextual claim} jointly expressed by the image--text pair. Yet most deepfake detectors, engineered to detect pixel-level forgeries, do not account for claim-level meaning, despite their growing integration in automated fact-checking (AFC) pipelines. This raises a central scientific and practical question: \textit{Do pixel-level detectors contribute useful signal for verifying image--text claims, or do they instead introduce misleading authenticity priors that undermine evidence-based reasoning?}
We provide the first systematic analysis of deepfake detectors in the context of multimodal misinformation detection. Using two complementary benchmarks such as MMFakeBench and DGM4, we evaluate: \textbf{(i)} state-of-the-art image-only deepfake detectors, \textbf{(ii)} an evidence-driven fact-checking system that performs tool-guided retrieval via Monte Carlo Tree Search (MCTS) and engages in deliberative inference through Multi-Agent Debate (MAD), and \textbf{(iii)} a hybrid fact-checking system that injects detector outputs as auxiliary evidence.
Results across both benchmark datasets show that deepfake detectors offer limited standalone value, achieving F1 scores in the range of 0.26–0.53 on MMFakeBench and 0.33–0.49 on DGM4, and that incorporating their predictions into fact-checking pipelines consistently \emph{reduces} performance by 0.04–0.08 F1 due to non-causal authenticity assumptions. In contrast, the evidence-centric fact-checking system achieves the highest performance, reaching F1 scores of approximately 0.81 on MMFakeBench and 0.55 on DGM4.
Overall, our findings demonstrate that multimodal claim verification is driven primarily by semantic understanding and external evidence, and that pixel-level artifact signals do not reliably enhance reasoning over real-world image--text misinformation.

\end{abstract}





\end{frontmatter}



\section{Introduction}

The rapid progress of generative AI has increasingly blurred the perceptual boundary between synthetic and authentic imagery, making it possible to generate, alter, and repurpose photorealistic visuals at an unprecedented scale \cite{a1}. However, growing empirical evidence shows that the principal harms of online misinformation stem not merely from the pixels of an image, but from the \emph{semantic narrative} that the image is mobilised to assert \cite{a2,a3,a4}. Modern misinformation is fundamentally multimodal: images are paired with captions, headlines, or short textual claims that jointly construct persuasive narratives, regardless of whether the underlying visual is synthetic or real.
In practice, such deception most often manifests as semantic and contextual misalignment, where \emph{contextual misalignment} involves authentic images reused with incorrect who/when/where framing, and \emph{semantic misalignment} involves a mismatch between what the image depicts and what the text asserts [3, 5] including AI-generated images crafted to support fabricated accounts or benign visuals repurposed to reinforce misleading interpretations. During the COVID-19 pandemic, for example, miscaptioned images such as “doctors collapsing after vaccination” circulated widely despite originating from unrelated incidents \cite{a6,a7}. Similar patterns appear in geopolitical misinformation, where AI-generated images are repeatedly shared as supposed evidence of troop movements in particular cities \cite{a5}. 
\textit{Across all scenarios, a clear pattern emerges: in practical settings, the damaging unit of misinformation is not the image alone, but the \emph{image–text claim} that the image is leveraged to reinforce}.

Current multimodal misinformation frequently mixes genuine photographs, recycled imagery, and AI-generated content, creating situations where pixel-level authenticity intersects with semantic distortion. Datasets such as \textsc{MMFakeBench} and \textsc{DGM$^{4}$} reflect this mix, containing both claim-level manipulations and subsets of synthetic or edited images. As deepfake imagery becomes increasingly common, it remains important to determine whether image-only detectors provide a useful signal for downstream verification.

\begin{figure}[t]
\centering
\includegraphics[width=0.8\textwidth]{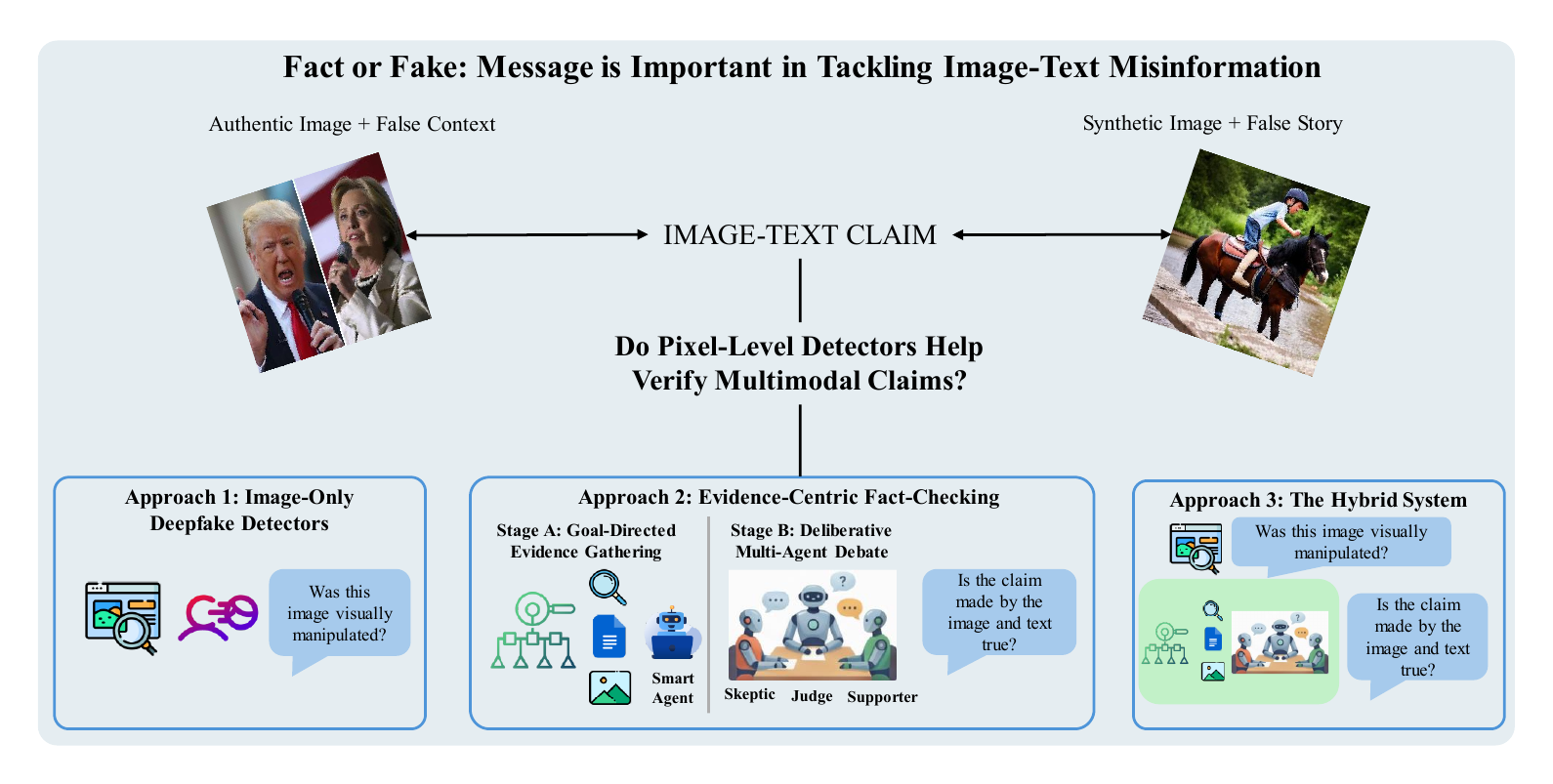}
\caption{Multimodal misinformation often uses authentic or synthetic images paired with false narratives. We compare (1) image-only deepfake detectors, (2) an evidence-centric fact-checking system, and (3) a hybrid that incorporates detector outputs, to ask whether pixel-level authenticity cues help verify image–text claims.}
\label{fig:0}
\end{figure}

At the same time, traditional image-based deepfake detectors were developed to capture visual artifacts such as frequency noise, blending seams, or generator fingerprints, achieving strong results on curated forensics benchmarks \cite{a8,a9}. These detectors generalize poorly in real-world multimodal settings where images may be authentic but contextually misleading, or synthetic yet semantically plausible \cite{a10}. Insights from cognitive science reinforce this limitation: authentic but miscaptioned images substantially increase perceived credibility \cite{a17}, narrative framing overrides visual fidelity \cite{a18,a3}, and humans rely heavily on textual context even when images visibly contradict it \cite{a19}. These findings highlight that \emph{the epistemic unit of multimodal misinformation is the image–text claim, not the image alone}. Nevertheless, deepfake detectors are widely deployed in moderation pipelines and increasingly used as auxiliary signals in automated fact-checking systems, raising an important open question: whether such artifact-centric cues meaningfully support claim verification or instead introduce misleading priors. This motivates our empirical evaluation of detectors both in isolation and integrated within an evidence-driven fact-checking pipeline.

Building on these insights, we conduct the first systematic study evaluating whether traditional pixel-level deepfake detectors offer any measurable utility for multimodal misinformation detection. As shown in Figure \ref{fig:0}, we examine three system families: \textbf{(i)} image-only deepfake detectors, \textbf{(ii)} an evidence-driven multimodal fact-checking system, and \textbf{(iii)} a hybrid variant that incorporates detector outputs as auxiliary cues. This evaluation is guided by the following research question:

\textbf{RQ}: Do image-based deepfake detectors provide any meaningful value for multimodal misinformation verification either as standalone predictors or as auxiliary signals within an evidence-centric fact-checking system, or do their artifact-based predictions introduce misleading priors that ultimately degrade claim-level reasoning?

To answer this question, we used two multimodal misinformation benchmarks datasets such as \textbf{MMFakeBench} and \textbf{DGM$^{4}$}, which together span diverse narrative themes, manipulation styles, and real-world misinformation patterns. These datasets allow us to assess both semantic/contextual distortion and visually manipulated content, providing a comprehensive testbed for understanding whether pixel-level authenticity signals transfer to claim-level verification.

The remainder of this paper is organized as follows. Section~2 reviews related work in multimodal misinformation detection, deepfake forensics, and automated fact-checking. Section~3 introduces our evidence-centric fact-checking framework and details the integration of deepfake detector outputs in the hybrid setting. Section~4 presents the experimental analysis. Section 6 discuss the case study of evidence-centric and hybrid fact-checking system. Finally, Section~5 concludes the paper.

\section{Related Work}

\subsection{Image-Based Deepfake Detection}
Image-based deepfake detection has long been the dominant paradigm for assessing visual authenticity. Early approaches relied on hand-crafted physiological or statistical cues such as eye-blink irregularities, color filter array inconsistencies, and illumination artifacts, while more recent deep learning methods—including transformer-based and frequency-domain models—learn latent generator fingerprints directly from data \cite{a8,a9}. Large-scale benchmarks such as FaceForensics++ \cite{a20}, Celeb-DF \cite{a21}, and DFDC \cite{a22} have standardized evaluation protocols and driven steady progress. Nevertheless, extensive cross-dataset studies consistently show that many detectors overfit to generator-specific artifacts and suffer substantial performance degradation under unseen synthesis methods, heavy compression, and platform-specific distortions \cite{a10}. To address these limitations, Hasanaath et al. proposed frequency-enhanced self-blended images, combining self-blending augmentation with discrete wavelet transform features to improve cross-dataset generalization \cite{a33}. Huang et al. introduced a generalized detection framework that adapts frozen foundation models such as CLIP via lightweight side-network adapters and released the DiGEN dataset to better capture diverse generative sources and unseen forgeries \cite{a34}. Extending beyond binary detection, Huang et al. further proposed SIDA, a large multimodal framework that jointly performs social-media image deepfake detection, tamper localization, and natural language explanation \cite{a35}. Complementarily, Kroiß et al. show that transfer-learned CNN backbones can remain strong baselines for face deepfake detection when trained on diverse forgeries such as DFFD \cite{a36}. In parallel, Arshed et al. reformulated deepfake detection as a multiclass problem and demonstrated that patch-wise Vision Transformer models can effectively distinguish between GAN, diffusion, and hybrid-generated faces by leveraging global contextual representations, outperforming conventional CNN baselines \cite{a37}.

Crucially, these detectors answer the question \emph{“Was this image visually manipulated?”} rather than \emph{“Is the associated claim true?”}. They reliably flag synthetic faces but remain insensitive to contextual falsity, such as authentic photos paired with misleading captions. Because most real-world misinformation arises from contextual and semantic distortion rather than pixel-level tampering, image-only detectors are misaligned with claim-level verification. This motivates part of our research question: do artifact-level signals from deepfake detectors provide any value for multimodal truth assessment, or do they introduce noise when authenticity and veracity diverge?

\subsection{Multimodal Misinformation Detection}



Multimodal misinformation detection systems have traditionally approached the task by assessing how semantically consistent an image–text pair appears, using fusion models that learn joint embeddings through CNN- or Transformer-based encoders. Representative systems such as ViLBERT, CLIP, and subsequent cross-modal attention models classify misinformation by measuring feature alignment across modalities \cite{a3,a5}. While such approaches outperform unimodal baselines, they remain fundamentally correlation-driven: an authentic image paired with a persuasive but false caption may still appear semantically coherent in the learned embedding space. To move beyond surface-level coherence, out-of-context (OOC) detection methods such as COSMOS \cite{a11}, CLIP-based entailment models \cite{a23}, and semantic inconsistency approaches \cite{a24} explicitly target mismatches between visual content and textual claims. Building on this direction, Qi et al. introduced SNIFFER, a multimodal large language model for explainable OOC detection that combines instruction tuning with retrieval-augmented reasoning over internal and external evidence \cite{a38}. Complementary efforts improve fusion itself: CAF-ODNN replaces naive concatenation with complementary bidirectional attention between post text and image captions, explicitly modeling subtle cross-modal discrepancies and improving detection performance across benchmarks \cite{a39}.

Recent advances increasingly shift toward evidence-grounded and large vision–language model-based verification. Tahmasebi et al. proposed an LVLM-based multimodal misinformation detection pipeline that performs unsupervised multimodal evidence retrieval followed by zero-shot fact verification, achieving strong cross-dataset generalization without task-specific fine-tuning \cite{a40}. Yan et al. presented TRUST-VL, a unified and explainable vision–language model trained with structured fact-checking chains to jointly address textual, visual, and cross-modal distortions, improving robustness and interpretability through evidence-grounded reasoning \cite{a41}. Extending toward holistic analysis, Xu et al. introduced MDAM3, a comprehensive framework for multitype multimodal misinformation that integrates internal detectors for AI-generated content and cross-modal inconsistencies with external web-based evidence retrieval and LVLM-driven explanation \cite{a42}. Despite these advances, existing methods largely frame misinformation detection as a single-stage verdict prediction problem and do not explicitly analyze when visual authenticity cues are informative or misleading for claim-level truth assessment. Collectively, this literature reveals that semantic coherence, contradiction awareness, improved fusion, or even evidence grounding alone does not guarantee factual correctness. Our work builds on these insights by replacing correlation-driven fusion with structured, evidence-seeking verification and by systematically examining the role of visual manipulation signals within a claim-centric multimodal fact-checking framework.

\subsection{Automated Fact-Checking for Multimodal Misinformation}


Automated fact-checking (AFC) reconceptualizes misinformation detection as a structured reasoning pipeline comprising claim parsing, evidence retrieval, and veracity estimation \cite{a12}. Recent multimodal AFC systems explicitly operationalize this framework. DEFAME \cite{a13} coordinates specialized retrieval agents for text and vision; LRQ-Fact \cite{a14} decomposes complex claims into sub-queries; CRAVE \cite{a15} clusters retrieved documents into narrative units; and Holmes \cite{a16} integrates retrieval-augmented LLM scoring. Recent work further explores tightly coupled multimodal AFC pipelines, such as Ahmed et al. combine CNN-based image validation with Transformer-based textual reasoning and cross-modal fusion to improve robustness over unimodal baselines \cite{a43}. Along similar lines, Kakizaki et al. propose MAFT, which textualizes multimodal inputs (image/video captioning and speech recognition), extracts claims, retrieves external evidence via web APIs, and generates interpretable fact-checking reports using LLMs—while also converting deepfake detector outputs into textual evidence for downstream reasoning \cite{a44}. More recently, Yang et al. introduce RAMA, a retrieval-augmented multi-agent framework that refines web queries for multimodal claims, aggregates cross-verified evidence, and ensembles multiple LLM judges to improve robustness and interpretability \cite{a45}. In addition, Kangur et al. propose MultiReflect, a multimodal self-reflective RAG-based AFC pipeline that iteratively retrieves, filters, ranks, and verifies image–text evidence from the internet, improving multimodal fact-checking accuracy \cite{a46}. Complementing retrieval-centric AFC, Chen et al. propose a causal-intervention and counterfactual multimodal fact-checking method that identifies causally relevant regions in retrieved visual evidence and suppresses shortcut unimodal reasoning via counterfactual decomposition, improving generalization on OOC benchmarks \cite{a47}. Finally, Zhang et al. introduce ESCNet and the large-scale CMFC dataset, enhancing multimodal fact-checking by jointly modeling stance reasoning between claims and retrieved evidence and incorporating knowledge-graph–grounded entity relations for stronger semantic and knowledge-level verification \cite{a48}. Other work emphasizes evidence alignment and explainability, showing that grounding both image and text in external knowledge improves multimodal verification \cite{a25,a26}.

Despite these advances, two key gaps remain. \textbf{First}, no prior work has systematically compared AFC pipelines with state-of-the-art image-based detectors on identical multimodal benchmarks, leaving open whether artifact-level cues contribute any measurable benefit. \textbf{Second}, the interaction between deepfake detectors and reasoning agents is poorly understood: detector outputs might enhance robustness or, alternatively, introduce misleading priors when images are authentic but miscaptioned. Addressing these open questions forms the empirical foundation of our study.

\begin{figure}[!htbp]
\centering
\includegraphics[width=0.9\textwidth]{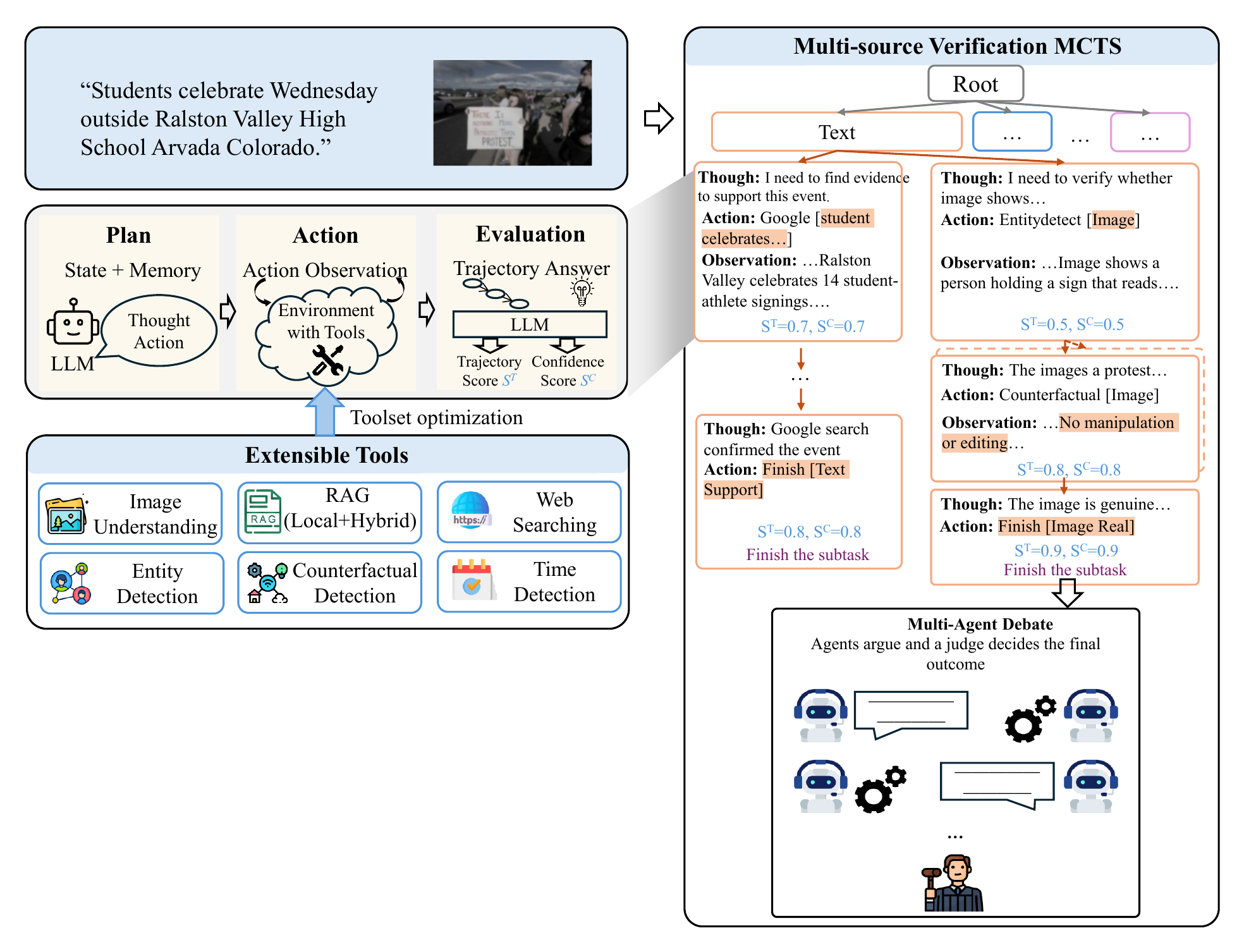}
\caption{Overview of the proposed evidence-centric fact-checking framework. Given an image--text claim, the system performs verification in two stages. \textbf{(1) Multi-source verification via MCTS:} a multimodal LLM plans tool calls across an extensible toolkit (e.g., web/RAG search, image understanding, counterfactual/VQA, entity and time detection), executes these actions within an environment, and evaluates each trajectory using utility and confidence scores to generate text-side and image-side stances with their corresponding evidence packs. \textbf{(2) Multi-Agent Debate (MAD):} skeptic and supporter agents deliberate over the collected evidence, while a neutral Judge LLM aggregates their arguments to issue the final verdict.}
\label{fig:1}
\end{figure}
\section{Method}
This section introduces our evidence-centric fact-checking method, a two-stage pipeline combining goal-directed evidence acquisition with deliberative adjudication, and then describes the image-based deepfake detectors used as baselines for comparison.

\subsection{Fact-Checking Method}

Our goal is to determine the truthfulness of an image--text claim using targeted tool use and structured debate-driven reasoning. As shown in Figure~\ref{fig:1}, a multimodal LLM serves as the central controller: it plans actions, invokes tools, interprets observations, and maintains memory throughout verification. The LLM selects from an extensible toolset depending on the active modality and subtask requirements, enabling unified reasoning across heterogeneous evidence sources.

To efficiently explore potential tool-use trajectories, we adopt a MCTS strategy inspired by T2Agent \cite{a27}. T2Agent formulates multimodal misinformation detection using an MCTS framework in which each node is evaluated by a dual reward function, where both the reasoning trajectory score and the confidence score are LLM-estimated and directly combined to guide node selection, backpropagation, and final decision aggregation across forgery-source subtasks. In contrast, our method explicitly disentangles trajectory utility from evidential confidence. The trajectory score $S_T$ is defined as a bounded structural measure that quantifies the quality of evidence acquisition, including progress, tool diversity, non-redundancy, and coherence of the action observation sequence, independent of claim veracity, while the confidence score is reserved exclusively for estimating the strength of factual support or refutation provided by the accumulated evidence for a given subtask. Rather than fusing $S_T$ and $S_C$ into a single decision signal, we use their product only as a search termination criterion, ensuring that high-confidence decisions are reached through sufficiently rich and non-degenerate evidence trajectories. Final claim verification is then performed by a MAD module, in which competing verifier agents reason adversarially over the structured evidence sets produced by MCTS, and a neutral judge aggregates their arguments into a final verdict. This separation of (i) trajectory optimization, (ii) confidence estimation, and (iii) deliberative decision making yields a more controlled and interpretable verification process.

Thus, multimodal claim verification is decomposed into two coordinated stages:

\begin{itemize}
    \item \textbf{Stage 1 – Goal-Directed Evidence Acquisition.}  
    An MCTS-guided multimodal LLM interacts with a suite of tools to obtain diverse, relevant, and reliable evidence for both text- and image-level subtasks.
    
    \item \textbf{Stage 2 – Deliberative Decision Making.}  
    The outputs of Stage~1 are adjudicated through MAD, producing a final, evidence-grounded decision.
\end{itemize}

\subsubsection{Goal-Directed Evidence Acquisition}

Evidence acquisition is framed as an LLM-controlled, goal-directed tree search over an extensible toolset. The system constructs a search tree using the classic \emph{selection} $\rightarrow$ \emph{expansion} $\rightarrow$ \emph{simulation} $\rightarrow$ \emph{backpropagation} cycle, where subtask-specific roots correspond to text and image verification.

At each node, the LLM proposes the next action based on the current state and trajectory, selects an appropriate tool, and processes the resulting observation as evidence for subtask decision making.

\paragraph{Search Structure}

The search tree captures successive reasoning trajectories guided by the LLM.  
The root corresponds to the overall verification goal for a claim $(I, T)$; its children represent the initial subtask roots $k \in \{\text{text},\ \text{image}\}$.

Each node stores a state
\begin{equation}
    s = \langle I, T, E, M, k \rangle ,
\end{equation}
where $E$ is the accumulated evidence, $M$ is the short-term memory of recent actions and observations, and $k$ is the active subtask.  
A node $n$ represents the pair $(s, \tau)$, where the trajectory
\[
    \tau = (a_{1:t}, o_{1:t})
\]
records all actions $a_i$ and observations $o_i$ along the path.

From state $s$, the planner proposes an action $a = \text{(tool, args)}$, executed in the environment to produce an observation $o$. This observation is normalized into an evidence atom
\begin{equation}
    e = \langle \text{modality} \in \{\text{text}, \text{image}\},\ \text{content},\ \text{source},\ \text{timestamp} \rangle .
\end{equation}
Each atom is appended to $E$, incrementally forming the evidence set used for stance estimation and coverage tracking.

\paragraph{Trajectory Evaluation}

Every partial trajectory  
\(
    \tau = (a_{1:t}, o_{1:t})
\)
is evaluated using two complementary bounded scores:

\begin{itemize}
    \item \textbf{Trajectory Utility} \( S^{T}(\tau) \in [0,1] \): assesses structural quality, coherence, and exploration diversity.

    \item \textbf{Confidence} \( S^{C}(\tau) \in [0,1] \): measures how strongly the collected observations support the current subtask label.
\end{itemize}

The final node value used during search is
\begin{equation}
    V(n) = \lambda S^{T}(\tau_n) + (1 - \lambda) S^{C}(\tau_n).
\end{equation}

\noindent\textbf{(a) Trajectory Utility \(S^{T}\).}  
This score rewards progressive, diverse, and coherent evidence acquisition. For a trajectory of length $L$, we compute:

\begin{itemize}
    \item \textbf{Progress:}
    \[
        \mathrm{prog}(L) = \frac{1}{1 + \exp(-(L-2))}.
    \]

    \item \textbf{Tool Diversity:}
    \[
        \mathrm{unique\_ratio}
        = \frac{|\{\text{tools}\}|}{\max(1, L)}.
    \]

    \item \textbf{Non-redundancy:}
    \[
        \mathrm{rep\_ratio}
        = \frac{\#\{\,i > 1 : \mathrm{tools}[i] = \mathrm{tools}[i-1]\}}{\max(1, L-1)},
    \qquad
        \mathrm{non\_repeat} = 1 - \mathrm{rep\_ratio}.
    \]
\end{itemize}

These are averaged into a structural base score:
\begin{equation}
    \mathrm{base}\,S^{T}(\tau)
    =
    \tfrac{1}{3}\left(
        \mathrm{prog}
        + \mathrm{unique\_ratio}
        + \mathrm{non\_repeat}
    \right).
\end{equation}

A compact LLM grader then assigns a coherence score  
\(
    \mathrm{llm}S^{T}(\tau) \in [0,1].
\)  
The final utility is:
\begin{equation}
    S^{T}(\tau)
    =
    \mathrm{clip}\left(
        \tfrac{1}{2}\,\mathrm{base}\,S^{T}(\tau)
        +
        \tfrac{1}{2}\,\mathrm{llm}S^{T}(\tau),
        0, 1
    \right).
\end{equation}

\noindent\textbf{(b) Confidence \(S^{C}\).}  
A grader LLM scores the reliability of the current evidence on a 1--10 scale:
\begin{equation}
    S^{C}(\tau) = \frac{s}{10}.
\end{equation}

\paragraph{Search Algorithm}

The MCTS process follows:

\begin{itemize}
    \item \textbf{Selection:}
    \[
        \mathrm{UCT}(n,a)
        =
        \frac{\widehat{Q}(n,a)}{N(n,a)+1}
        +
        C\sqrt{
            \frac{\ln(N(n)+1)}{N(n,a)+1}
        }.
    \]

    \item \textbf{Expansion:} the best $(n,a)$ pair is executed; a child node is created.

    \item \textbf{Simulation:} a lightweight rollout collects provisional scores $(S^{T}, S^{C})$.

    \item \textbf{Backpropagation:} the combined value \(V\) is propagated upward.
\end{itemize}

Each subtask runs under a fixed step budget \(B_k\) and terminates early when its best trajectory \(\tau_k^{*}\) satisfies:
\begin{equation}
    S^{T}(\tau_k^{*}) \cdot S^{C}(\tau_k^{*}) \ge \theta_k.
\end{equation}

Branches yielding high-confidence falsity (e.g., explicit refutation or verified out-of-context provenance) are pruned, and their unused budget is reallocated. This reflects the monotonicity that a claim cannot be \textsc{Real} if any component is confidently \textsc{Fake}.

Stage~1 returns structured summaries:
\begin{equation}
    (z_T, E^{\text{text}}, \tau_T^{*}), 
    \qquad
    (z_I, E^{\text{image}}, \tau_I^{*}),
\end{equation}
where
\[
    z_T \in \{\text{TEXT\_REAL}, \text{TEXT\_FAKE}\}, 
    \qquad
    z_I \in \{\text{IMAGE\_REAL}, \text{IMAGE\_FAKE}\}.
\]

\begin{algorithm}[!htbp]
\caption{Goal-Directed Evidence Acquisition}
\label{alg:mcts_evidence}
\begin{algorithmic}[1]
\Require Claim $(I, T)$, subtasks $k \in \{\text{text}, \text{image}\}$,\\
\hspace{1.4em} tool set $\mathcal{A}$, step budgets $\{B_k\}$, thresholds $\{\theta_k\}$, mixing weight $\lambda \in [0,1]$
\Ensure For each subtask $k$: label $z_k$, evidence $E_k$, best trajectory $\tau_k^\ast$

\For{each subtask $k \in \{\text{text}, \text{image}\}$}
    \State Initialize evidence $E \gets \emptyset$, memory $M \gets \emptyset$
    \State Initialize state $s_0 \gets \langle I, T, E, M, k \rangle$
    \State Initialize trajectory $\tau_0 \gets ()$
    \State Create root node $n_0 \gets (s_0, \tau_0)$
    \State Initialize search tree $\mathcal{T} \gets \{n_0\}$
    \For{$b = 1$ to $B_k$}
        \Comment{MCTS iteration}
        \State $n \gets n_0$; $\text{path} \gets [n_0]$
        \While{$n$ is fully expanded \textbf{and} $n$ is not terminal}
            \State Select action $a^\ast = \arg\max_{a} \mathrm{UCT}(n,a)$
            \State $n \gets \text{Child}(n, a^\ast)$
            \State Append $n$ to \text{path}
        \EndWhile
        \If{$n$ is not terminal}
            \State $a \gets \Call{LLMPlanner}{n.\text{state}, n.\tau}$ 
            \Comment{$a = (\text{tool}, \text{args})$}
            \State $(s', \tau') \gets \Call{StepEnvironment}{n.\text{state}, n.\tau, a}$
            \State Create child node $n' \gets (s', \tau')$ and attach to $n$
            \State $\mathcal{T} \gets \mathcal{T} \cup \{n'\}$
            \State $n \gets n'$; append $n$ to \text{path}
        \EndIf
        \State $(S_T, S_C) \gets \Call{RolloutAndScore}{n}$
        \Comment{lightweight rollout}
        \State $V \gets \lambda S_T + (1 - \lambda) S_C$
        \ForAll{$m \in \text{path}$}
            \State $\Call{UpdateStats}{m, V}$ 
            \Comment{update $Q, N$ for actions leading to $m$}
        \EndFor
        \State $n^\ast \gets \arg\max_{m \in \mathcal{T}} V(m)$
        \State $(S_T^\ast, S_C^\ast) \gets \Call{RecomputeScores}{n^\ast}$
        \If{$S_T^\ast \cdot S_C^\ast \ge \theta_k$}
            \State \textbf{break}
        \EndIf
    \EndFor
    \State $n_k^\ast \gets \arg\max_{m \in \mathcal{T}} V(m)$
    \State $(s_k^\ast, \tau_k^\ast) \gets n_k^\ast$
    \State $E_k \gets s_k^\ast.E$
    \State $z_k \gets \Call{SubtaskLabel}{E_k, k}$ 
\EndFor
\State \Return $(z_T, E_{\text{text}}, \tau_T^\ast), (z_I, E_{\text{image}}, \tau_I^\ast)$
\end{algorithmic}
\end{algorithm}

Algorithm 1 summarizes our goal-directed evidence acquisition via MCTS. The algorithm takes as input a claim $c$, a set of subtasks $K=\{\textsc{Text},\textsc{Image}\}$, an extensible tool set $\mathcal{T}$, subtask-specific step budgets $\{B_k\}_{k\in K}$, termination thresholds $\{\tau_k\}_{k\in K}$, and a mixing weight $\lambda\in[0,1]$ for combining structural utility and confidence scores. For each subtask $k\in K$, the algorithm returns a discrete label $y_k$, an evidence set $\mathcal{E}_k$, and the best reasoning trajectory $\pi_k^\star$.

For each subtask $k$, we initialize an empty evidence buffer $\mathcal{E}\leftarrow\emptyset$ and a short-term memory $\mathcal{M}$ that tracks recent actions and observations. Together with the claim $c$ and the active subtask $k$, these define the initial state $s_0$. We initialize an empty trajectory $\pi\leftarrow[\ ]$, create the root node $n_0\!\leftarrow\!\textsc{Node}(s_0)$, and start the search tree with this root.

Each MCTS iteration (up to budget $B_k$) consists of the four standard phases: selection, expansion, simulation, and backpropagation. In the selection phase, we start from the root and repeatedly select actions according to a UCT-style exploration--exploitation rule. At each internal node, we choose the best action among its outgoing edges, transition to the corresponding child node, and record visited nodes along this path in a list $\textit{path}$. This continues until reaching a node that is terminal or not yet fully expanded.

In the expansion phase, if the current node is non-terminal, we invoke an LLM-based planner $\textsc{LLMPlanner}(n.\textit{state},k,\mathcal{T})$ to propose the next action (tool, args). The environment step $\textsc{StepEnvironment}(n.\textit{state}, a)$ executes this tool call and returns an updated state with new evidence and memory, together with an extended trajectory that appends the new action--observation pair. We then create a new child node, attach it to its parent, insert it into the tree, and extend the current path.

In the simulation phase, starting from the expanded node, we perform a lightweight rollout via $\textsc{RolloutAndScore}(\cdot)$. This routine stochastically continues the trajectory for a small number of steps using inexpensive heuristics, then evaluates the resulting trajectory with two bounded scores: (i) a structural utility $U\in[0,1]$ capturing progress, diversity, and coherence of evidence acquisition; and (ii) a confidence score $S\in[0,1]$ capturing how strongly the accumulated evidence supports a subtask decision. The overall node value is computed as a convex combination $V=\lambda U+(1-\lambda)S$.

During backpropagation, the value $V$ is propagated back along all nodes in $\textit{path}$. For each node on the path, $\textsc{UpdateStats}(\cdot)$ updates visit counts and value estimates associated with the actions that led to the next node. These statistics are subsequently used by the UCT rule during future selection steps.

After each iteration, we identify the current best node in the tree according to its value estimate and recompute its scores via $\textsc{RecomputeScores}(\cdot)$ to obtain refined structural utility and confidence $(U,S)$. If their product exceeds the subtask-specific threshold, i.e., $U\cdot S \ge \tau_k$, the search terminates early, reflecting that the system has acquired sufficiently strong and coherent evidence for a stable decision.

At the end of the search (either by budget exhaustion or early stopping), we select the globally best node for subtask $k$ from the tree. Its state contains the final evidence buffer $\mathcal{E}_k$, and its trajectory encodes the full reasoning path $\pi_k^\star$ used to collect this evidence. A subtask-specific decision function $\textsc{SubtaskLabel}(\mathcal{E}_k,k)$ maps the evidence to the final label $y_k$ (e.g., $\{\textsc{Text-Real},\textsc{Text-Fake}\}$ for text and $\{\textsc{Image-Real},\textsc{Image-Fake}\}$ for image).

\subsubsection{Multi-Agent Debate (MAD) for Decision Making}

MAD produces the final decision using Stage~1 evidence.  
For each input $(I, T)$, the MCTS stage provides modality-specific labels, calibrated confidence scores, and compact evidence packs. MAD transforms these summaries into a consensus through controlled adversarial dialogue.

\paragraph{Label Space and Initialization}

The overall label space is:
\[
    Y_4 = \{\text{REAL},\ \text{TEXT\_FAKE},\ \text{IMAGE\_FAKE},\ \text{BOTH\_FAKE}\}.
\]

Two agents assume complementary roles: Agent~A acts as a skeptic (falsification-first), and Agent~B as a supporter (verification-first).

\paragraph{Round-Based Debate Process}

For \(R\) rounds, each agent \(j\in\{A,B\}\) outputs:
\[
    y_{r}^{(j)} \in Y,\qquad 
    c_{r}^{(j)} \in [0,1],\qquad 
    \rho_{r}^{(j)} ,
\]
where \(y_{r}^{(j)}\) is the proposed label, \(c_{r}^{(j)}\) is its confidence, and \(\rho_{r}^{(j)}\) is a rationale citing evidence from Stage~1.

Define the novelty factor:
\[
    \kappa_{r}^{(j)} =
    \begin{cases}
        1, & \text{if } C_{r}^{(j)} \setminus C_{<r}^{(j)} \neq \emptyset, \\[4pt]
        \pi, & \text{otherwise},
    \end{cases}
\]
where $\pi\in(0,1)$ penalizes repeated evidence use.

\paragraph{Judgment and Final Verdict}

After the final round, a neutral Judge LLM examines the full transcript and Stage~1 summaries and outputs:
\[
    y_{J} \in Y, \qquad c_{J} \in [0,1].
\]
This constitutes the system’s final claim-level decision.

Algorithm~2 applies a multi-agent debate (MAD) layer on top of the Stage--1 evidence bundle, which aggregates modality-specific subtasks $k\in K$ with labels $y_k\in Y_4$, confidences $s_k\in[0,1]$, and cited evidence sets $\mathcal{C}_k$. Two LLM agents debate the claim: a skeptic $A$ (falsification-first) and a supporter $B$ (verification-first). Across up to $R$ rounds, they alternately propose labels $\ell_A,\ell_B\in Y_4$ with confidences $c_A,c_B\in[0,1]$, rationales $\rho_A,\rho_B$, and citation sets $\mathcal{C}^{(A)}_r,\mathcal{C}^{(B)}_r$. Each agent maintains a citation history $\mathcal{C}^{(j)}_{<r}$ and receives a novelty factor (1 for new evidence, $\pi$<1 for fully reused citations) which down-weights its effective confidence $\bar{c}^{(j)}_r = \kappa^{(j)}_r \, c^{(j)}_r$. All turns are logged in a global transcript $\mathcal{T}$, and the debate may terminate early if both agents agree on the same label.

In the judge phase, a separate LLM judge is provided with the full transcript $\mathcal{T}$ and the final turns. It outputs a candidate label $\ell_J\in Y_4$, confidence $c_J\in[0,1]$, and an override flag $o_J\in\{0,1\}$. If $c_J$ exceeds a threshold $\tau_J$, the MAD outcome is adopted as the final decision $y^\star=\ell_J$; otherwise, the system falls back to a simple fusion of Stage--1 labels and confidences $\{(y_k,s_k)\}_{k\in K}$. The procedure also returns a diagnostic record summarizing whether MAD was trusted, the judge's choice, the number of rounds used, the final agent states, and the full transcript, enabling transparent analysis of how the debate influenced the final claim decision.

\begin{algorithm}[!htbp]
\caption{Multi-Agent Debate (MAD) for Final Decision}
\begin{algorithmic}[1]
\Require Stage--1 evidence bundle with modality-specific subtasks $k\in K$ (e.g., $K=\{\text{text},\text{image}\}$). 
\Ensure Final label $y^\star \in Y_4$.
\State \textbf{Initialize debate:} $\textit{transcript}\gets[\ ]$; $\textit{a\_last}\gets\bot$; $\textit{b\_last}\gets\bot$.
\State For each agent $j\in\{A,B\}$, define cumulative citation history $\mathcal{C}^{(j)}_{<1}\gets\emptyset$.
\State Agent roles: $A$ = skeptic (falsification-first), $B$ = supporter (verification-first).

\For{$r=1$ \textbf{to} $R$}
  \Comment{Round-based debate}

  \State \textbf{Agent A (Skeptic) turn:} provide $\{B,\ \textit{transcript},\ r,\ R,\ \text{role}=\text{skeptic}\}$.
  \State Receive structured output $(\ell_A,\ c_A,\ \rho_A,\ \mathcal{C}^{(A)}_{r})$ where
         $\ell_A\in Y_4$ is the proposed label, $c_A\in[0,1]$ its confidence,
         $\rho_A$ a rationale, and $\mathcal{C}^{(A)}_{r}\subseteq\mathcal{C}$ the cited evidence.
  \State Compute novelty factor
        $\kappa^{(A)}_{r} \gets
          \begin{cases}
            1, & \text{if }\mathcal{C}^{(A)}_{r}\setminus \mathcal{C}^{(A)}_{<r} \neq \emptyset\\
            \pi, & \text{otherwise}
          \end{cases}$
  \State Effective confidence $\tilde{c}_A \gets \kappa^{(A)}_{r}\cdot c_A$.
  \State Update history $\mathcal{C}^{(A)}_{<r+1} \gets \mathcal{C}^{(A)}_{<r}\cup \mathcal{C}^{(A)}_{r}$.
  \State Append $\langle A,\ r,\ \ell_A,\ \tilde{c}_A,\ \rho_A,\ \mathcal{C}^{(A)}_{r},\ \kappa^{(A)}_{r}\rangle$ to \textit{transcript};
         $\textit{a\_last}\gets(\ell_A,\tilde{c}_A,\rho_A,\mathcal{C}^{(A)}_{r})$.

  \State \textbf{Agent B (Supporter) turn:} provide $\{B,\ \textit{transcript},\ r,\ R,\ \text{role}=\text{supporter}\}$.
  \State Receive $(\ell_B,\ c_B,\ \rho_B,\ \mathcal{C}^{(B)}_{r})$.
  \State $\kappa^{(B)}_{r} \gets
          \begin{cases}
            1, & \text{if }\mathcal{C}^{(B)}_{r}\setminus \mathcal{C}^{(B)}_{<r} \neq \emptyset\\
            \pi, & \text{otherwise}
          \end{cases}$,
        \quad $\tilde{c}_B \gets \kappa^{(B)}_{r}\cdot c_B$.
  \State $\mathcal{C}^{(B)}_{<r+1} \gets \mathcal{C}^{(B)}_{<r}\cup \mathcal{C}^{(B)}_{r}$.
  \State Append $\langle B,\ r,\ \ell_B,\ \tilde{c}_B,\ \rho_B,\ \mathcal{C}^{(B)}_{r},\ \kappa^{(B)}_{r}\rangle$ to \textit{transcript};
         $\textit{b\_last}\gets(\ell_B,\tilde{c}_B,\rho_B,\mathcal{C}^{(B)}_{r})$.

  \If{\texttt{stop\_on\_consensus} \textbf{and} $\ell_A=\ell_B$}
    \State \textbf{break} \Comment{early stop when both agents agree on the label}
  \EndIf
\EndFor

\State \textbf{Judge phase:}
\State Provide Judge with $\{B,\ \textit{transcript},\ \textit{a\_last},\ \textit{b\_last}\}$.
\State Receive judge output $(\ell_J,\ c_J,\ o_J)$ with label $\ell_J\in Y_4$.

\If{$c_J \ge \texttt{judge\_min\_conf}$}
  \State $y^\star \gets \ell_J$;
\Else
  \State Obtain $y^\star$ from simple fusion of Stage--1 labels/confidences.
\EndIf
\end{algorithmic}
\end{algorithm}

\subsection{Image-Only Deepfake Detectors}

To contrast artifact-centric perception with evidence-centric reasoning, we evaluate five representative image-only detectors such as GRAMNet\cite{a28}, BNext-M \cite{a29}, NPR \cite{a30}, FatFormer \cite{a31}, and ProDet \cite{a32}, where each exemplifying a major design paradigm in visual forensics. All models operate solely on the image, without access to text or external evidence, and are evaluated under identical preprocessing, thresholding, and scoring protocols to ensure fairness and reproducibility.

\begin{itemize}

    \item \textbf{GRAMNet.}  
    GRAMNet models cross-layer Gram-matrix statistics to capture generator-specific texture and frequency correlations. It is representative of frequency-fingerprint approaches that achieve strong performance on curated benchmarks but degrade under unseen generators or heavy compression.  
    \textit{Justification:} Represents classical frequency-domain detectors.

    \item \textbf{BNext-M.}  
    BNext-M emphasizes physiological and structural consistency, learning illumination and facial micro-geometry cues via convolutional backbones. It performs well on portrait-based forgeries but struggles on contextual or non-face manipulations.  
    \textit{Justification:} Covers structure-aware models common in early deepfake detection.

    \item \textbf{NPR}  
    NPR leverages sensor-noise priors and patch-residual statistics to flag local inconsistencies between natural and synthesized regions. It generalizes moderately across generators but is easily disrupted by benign image transformations.  
    \textit{Justification:} Captures residual-based anomaly detectors.

    \item \textbf{FatFormer.}  
    FatFormer employs transformer encoders with artifact-aware objectives, modeling long-range dependencies and achieving state-of-the-art accuracy on several forensics benchmarks. However, it tends to over-flag stylistically atypical yet authentic images.  
    \textit{Justification:} Represents high-capacity transformer-based detectors common in recent leaderboards.

    \item \textbf{ProDet.}  
    ProDet focuses on provenance-agnostic cues such as camera-pipeline and compression irregularities, aiming to generalize across unknown generators but often trading precision for recall.  
    \textit{Justification:} Provides a manipulation agnostic baseline for general-purpose detection.

\end{itemize}

\subsection{Hybrid Fact-Checking System with Deepfake-Detector}

For the hybrid model used to address RQ2, we extend the fact-checking pipeline by incorporating traditional image-based deepfake detector predictions as additional image-side evidence. The detector’s binary authenticity verdict (\emph{real} or \emph{fake}) and its associated confidence score are first converted into a structured evidence unit, which is then injected into both stages of the fact-checking process.

During the MCTS-guided search, this detector-derived evidence unit is added to the initial evidence set $E$ available to the planner. The multimodal LLM may query or reference this unit when forming intermediate hypotheses about the image’s authenticity or when deciding whether further provenance, temporal, or semantic evidence is required. In this way, the detector’s output can influence the search trajectory and is treated as one evidence source among others, alongside retrieved items from web/RAG, image analysis, and other tools.

The same evidence unit is passed to the Multi-Agent Debate module as part of the shared evidence pack provided to the supporter and skeptic agents. Both agents may cite the authenticity cue when constructing arguments, and the Judge model considers the detector’s output when synthesizing the debate transcript and issuing the final decision. This integration strategy ensures that deepfake detectors function strictly as auxiliary evidence within the decision process, mirroring real-world conditions in which artifact-level authenticity cues shape users’ prior beliefs and judgments. The hybrid model thus provides a controlled setting for evaluating whether pixel-level forgery signals genuinely contribute to or instead interfere with evidence-centric multimodal reasoning.

\section{Results and Analysis}

\subsection{Benchmarks}

We evaluate our systems on two multimodal misinformation benchmarks: \textbf{MMFakeBench} and \textbf{DGM4}. Following established practice, we use the official \emph{test} split of MMFakeBench and the official \emph{validation} split of DGM4.

The datasets span diverse narrative domains (politics, crisis, health, entertainment), mixtures of authentic and synthetic imagery, and frequent contextual misalignment (authentic images paired with false captions). MMFakeBench emphasizes caption--image consistency with broad topical coverage, whereas DGM4 focuses on manipulation diversity and generator heterogeneity. Together, they capture both \textit{(i)} narrative/context falsity and \textit{(ii)} artifact diversity.

We evaluate all methods using \textbf{accuracy}, \textbf{precision}, \textbf{recall}, and \textbf{F1-score}.  
These metrics jointly measure overall correctness and the trade-off between false alarms and missed detections—both crucial for multimodal misinformation detection. Accuracy reflects global reliability, while precision, recall, and F1-score capture sensitivity to false claims and robustness against misclassifying true ones, providing a fair and interpretable basis for comparison.






\subsection{Quantitative Analysis}

To ensure clarity and direct comparability, we report results using a single unified table per dataset. Each table includes (i) five representative deepfake detectors, (ii) the hybrid detector-augmented fact-checking model, and (iii) the evidence-centric fact-checking system.  
This consolidated reporting eliminates ambiguity and allows strict side-by-side comparison under identical evaluation settings.

\subsubsection{MMFakeBench}

Table~\ref{tab:mmfakebench-unified} presents the unified comparison on MMFakeBench.  
Among the detectors, FatFormer achieves the highest performance (Acc = 0.79, F1 = 0.53), followed by BNext-M and NPR (F1 $\approx$ 0.53). GRAMNet attains reasonable accuracy (0.75) but weaker balanced metrics (F1 = 0.48), while ProDet performs poorly under claim-level supervision (F1 = 0.26). Overall, artifact-only cues plateau around F1 $\approx 0.53$, confirming that pixel-based signals do not transfer well to a benchmark dominated by caption-driven and out-of-context misinformation.

Integrating detector predictions into the fact-checking system \emph{reduces} performance (Acc = 0.76, F1 = 0.74). Although the hybrid model benefits from evidence-gathering modules, the injected pixel-level authenticity prior introduces non-causal noise particularly harmful when authentic images accompany deceptive captions. Our evidence-centric system achieves the strongest results (Acc = 0.82, F1 = 0.81), outperforming all baselines across all metrics.

Table~\ref{tab:mmfakebench-unified} highlights a clear trend:  
\textit{artifact-centric detectors plateau, hybridization degrades results, and evidence-driven reasoning performs best.}

\begin{figure}[!htbp]
\centering
\includegraphics[width=0.9\textwidth]{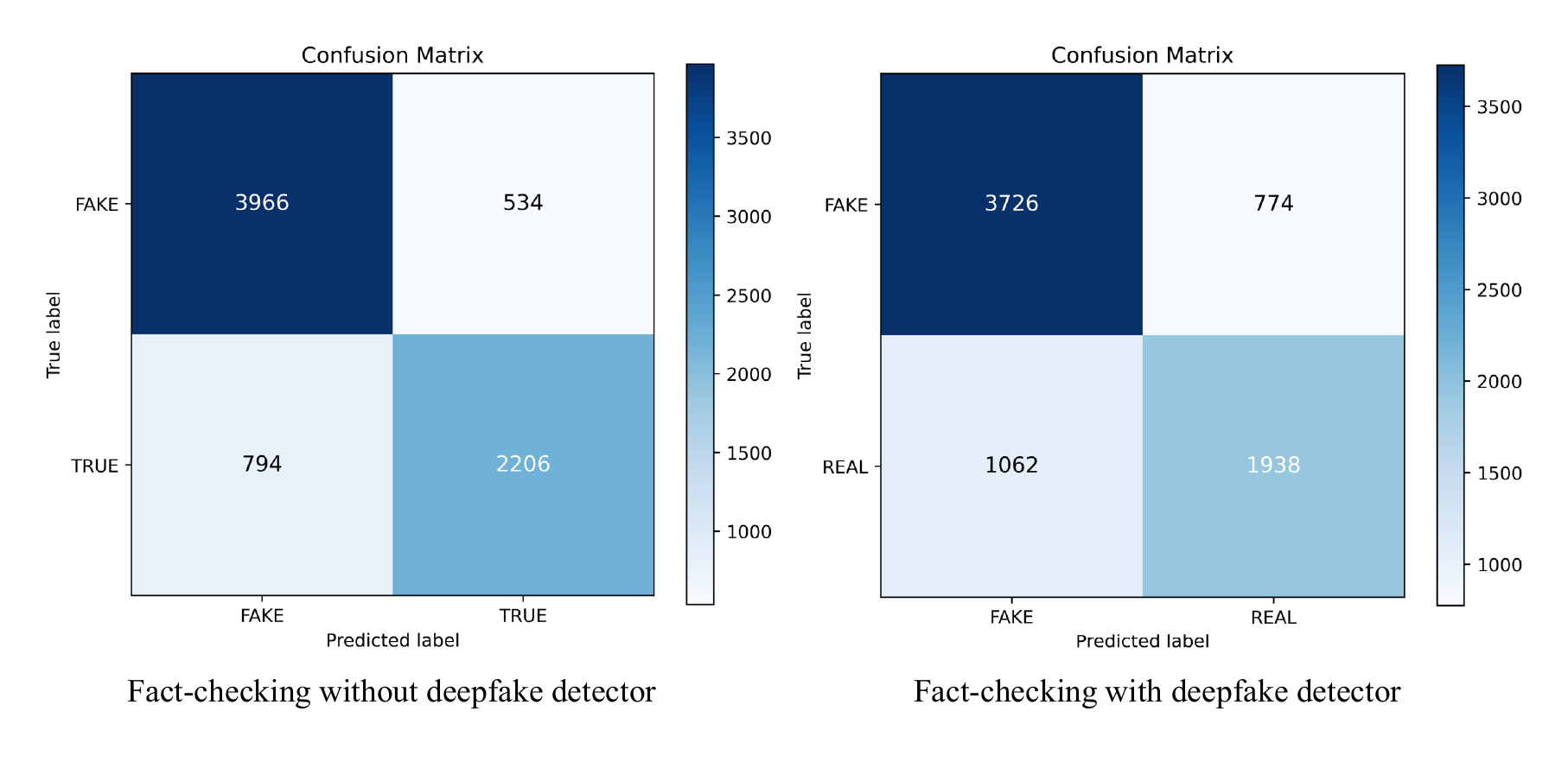}
\caption{The confusion matrix of the fact-checking with and without image-based deepfake detector on MMFakebench dataset.}
\label{fig:3}
\end{figure}

A deeper analysis explains why detector integration consistently harms performance which can be seen from Figure~\ref{fig:3}. Most MMFakeBench “fake’’ instances stem from textual distortions, repurposing, semantic contradiction, or temporal mismatch; only a minority contain visual manipulations detectable by pixel-level forensic models. Consequently, deepfake detectors tend to predict \emph{real} for many false claims. When these misleading priors are injected into the fact-checker, they conflict with retrieved provenance or semantic evidence. Confusion matrices confirm that the hybrid variant exhibits higher false-negative \emph{and} false-positive rates, indicating that detector signals suppress the fact-checker’s ability to flag false claims and simultaneously inflate errors on true claims. These results verify that on MMFakeBench, where misinformation is overwhelmingly semantic, pixel-level authenticity signals undermine, rather than enhance, multimodal verification.

\begin{table}[t]
\centering
\caption{Unified performance comparison on MMFakeBench across image-only deepfake detectors, the evidence-centric fact-checking system, and the hybrid (detector-integrated) variant.}
\label{tab:mmfakebench-unified}
\begin{tabular}{lcccc}
\toprule
\textbf{Method} & \textbf{Accuracy} & \textbf{Precision} & \textbf{Recall} & \textbf{F1-score} \\
\midrule
GRAMNet                           & 0.74 & 0.48 & 0.49 & 0.47 \\
BNext-M                           & 0.74 & 0.58 & 0.51 & 0.53 \\
NPR                               & 0.64 & 0.52 & 0.53 & 0.52 \\
FatFormer                         & 0.79 & 0.61 & 0.53 & 0.53 \\
ProDet                            & 0.26 & 0.54 & 0.52 & 0.25 \\
\midrule
Fact-checking + deepfake detector & 0.76 & 0.74 & 0.74 & 0.74 \\
Fact-checking (ours)              & \textbf{0.82} & \textbf{0.82} & \textbf{0.81} & \textbf{0.81} \\
\bottomrule
\end{tabular}
\end{table}

\subsubsection{DGM4}
Unified results for DGM4 are shown in Table~\ref{tab:dgm4-unified}.  
DGM4 is more heterogeneous, with substantial generator diversity and mixed manipulation styles. Among the detectors, ProDet achieves the best F1-score (0.49), outperforming FatFormer, GRAMNet, and NPR (F1 $\approx$ 0.33–0.35). Yet all detectors fall within a narrow accuracy band (0.46–0.52), again highlighting the limited transferability of artifact cues to claim-level tasks.

The hybrid fact-checking + detector model yields F1 = 0.51, slightly below our evidence-centric system. While it benefits from structured evidence acquisition, the detector’s artifact priors introduce ambiguity for cases involving authentic images paired with false captions or temporal inconsistencies. Our evidence-centric system achieves the highest performance overall (Acc = 0.55, F1 = 0.55). Its behavior reflects DGM4’s difficulty: the model expresses high precision for \textsc{FAKE} and high recall for \textsc{REAL}, consistent with the challenge of reasoning across unseen generators and diverse narrative structures.

\begin{figure}[!htbp]
\centering
\includegraphics[width=0.9\textwidth]{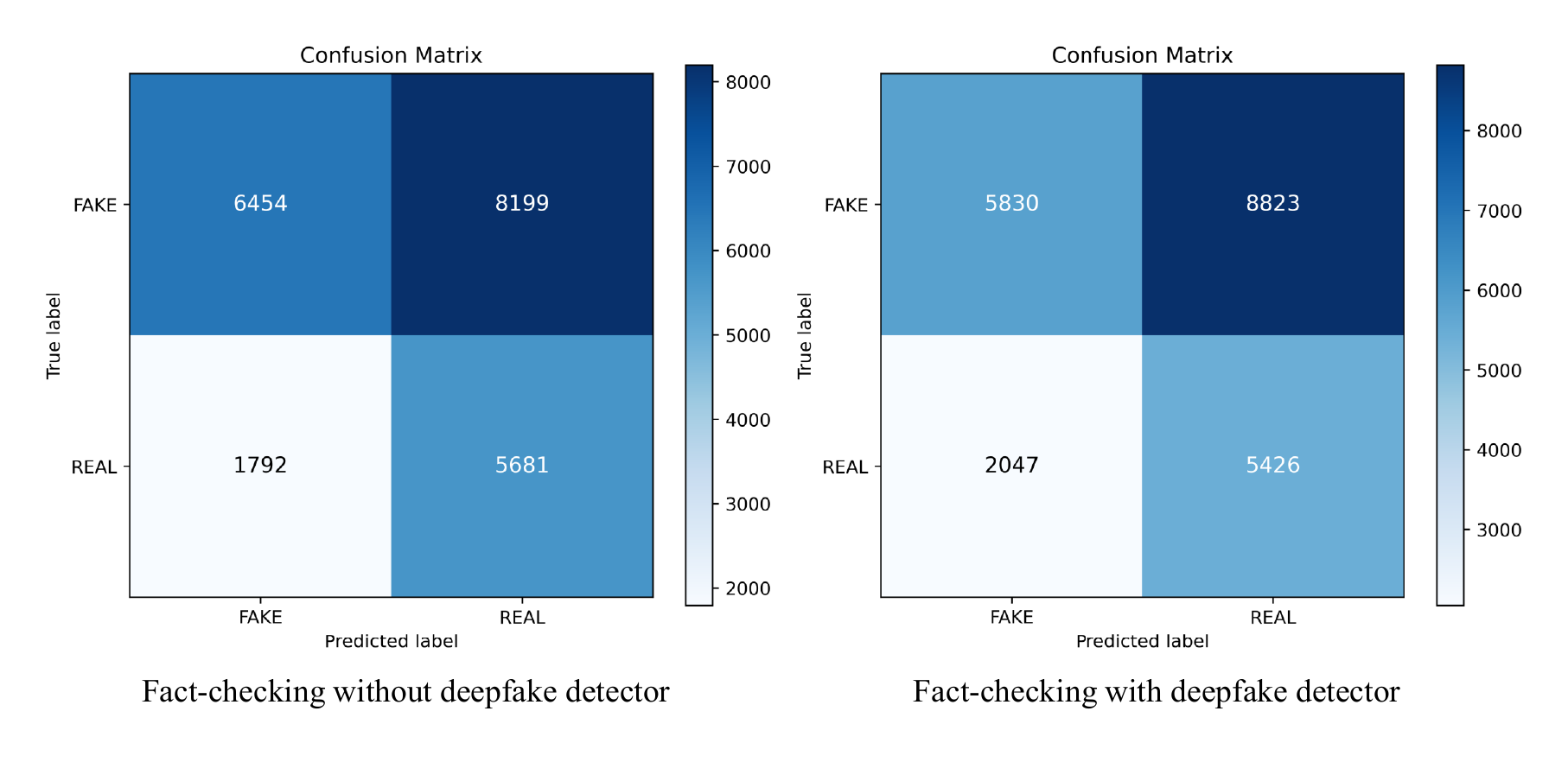}
\caption{Confusion matrices on the DGM$^{4}$ dataset for the evidence-centric fact-checking system and the hybrid system integrating a deepfake detector.}
\label{fig:4}
\end{figure}

A more granular analysis reveals why detector integration again fails to help. Although DGM4 includes genuine visual manipulations, such as face swaps (FS) and face attribute edits (FA), the manipulated regions are often tiny, subtle, or sentiment-driven. As a result, many fake claims remain visually plausible, and detectors frequently predict \emph{real} even for manipulated cases, which is evident from Figure~\ref{fig:4}. Furthermore, hybrid (face+text) manipulations constitute only a minority, weakening the correlation between pixel artifacts and claim-level falsity. Injecting these weak priors into the fact-checker introduces systematic bias, diminishing the influence of provenance, entailment, or contradiction cues. Confusion-matrix analyses confirm that the hybrid model exhibits elevated misclassification rates across classes. Thus, even in a dataset with more visible manipulations, artifact cues remain unreliable for claim verification.

\begin{table}[h]
\centering
\caption{Unified performance comparison on DGM4 across image-only deepfake detectors, the evidence-centric fact-checking system, and the hybrid (detector-integrated) variant.}
\label{tab:dgm4-unified}
\begin{tabular}{lcccc}
\toprule
\textbf{Method} & \textbf{Accuracy} & \textbf{Precision} & \textbf{Recall} & \textbf{F1-score} \\
\midrule
GRAMNet                           & 0.46 & 0.51 & 0.50 & 0.35 \\
BNext-M                           & 0.46 & 0.49 & 0.49 & 0.37 \\
NPR                               & 0.45 & 0.46 & 0.49 & 0.33 \\
FatFormer                         & 0.46 & 0.60 & 0.50 & 0.33 \\
ProDet                            & 0.51 & 0.50 & 0.50 & 0.49 \\
\midrule
Fact-checking + deepfake detector & 0.51 & 0.56 & 0.55 & 0.51 \\
Fact-checking (ours)              & \textbf{0.55} & \textbf{0.60} & \textbf{0.61} & \textbf{0.55} \\
\bottomrule
\end{tabular}
\end{table}

\subsection{Qualitative Analysis}



\begin{figure}[!htbp]
\centering
\includegraphics[width=0.9\textwidth]{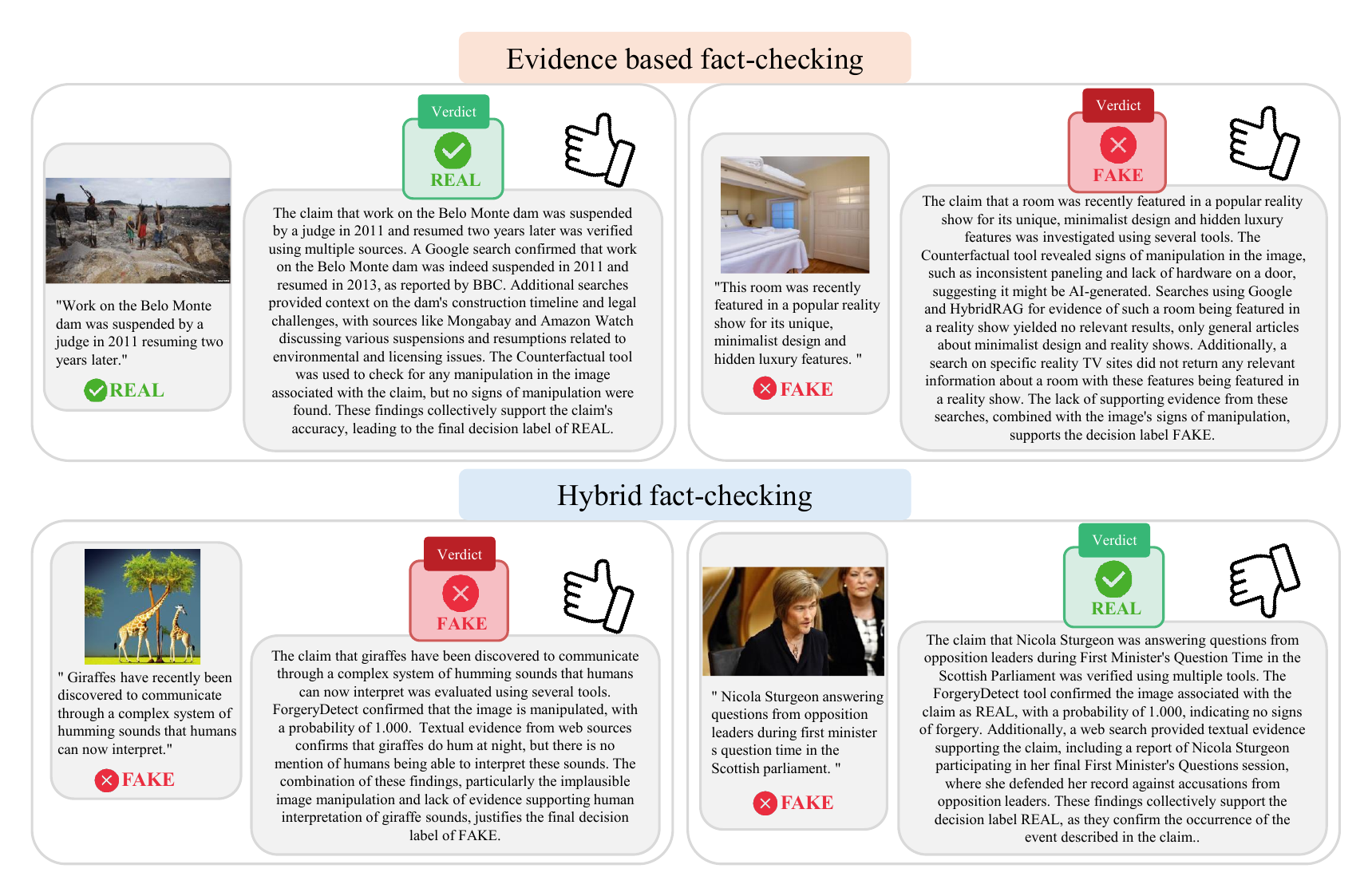}
\caption{Qualitative cases showing robust evidence-centric verification and a failure mode of hybrid fact-checking caused by misleading deepfake detector outputs.}
\label{fig:5}
\end{figure}

Figure~\ref{fig:5} shows qualitative examples comparing our evidence-based fact-checking system with the hybrid variant that also uses a deepfake detector signal. The top examples highlight how evidence-centric fact-checking can succeed even when the image itself is visually plausible. In the REAL case, the system finds supporting sources that match the key details in the caption (e.g., event context and timing), and the image content appears consistent with that context. In the FAKE case, the image looks ordinary and could fit many narratives, but external sources and provenance checks do not support the specific story asserted by the text. In both cases, the model treats the image as helpful context, while letting the final decision be driven by whether the claim can be substantiated.

The bottom row illustrates the contrasting behavior of the hybrid fact-checking system, which integrates deepfake detector outputs as auxiliary cues. In the left example, the hybrid model correctly labels a clearly synthetic image as FAKE because the forgery detector produces a confident IMAGE\_FAKE signal that aligns with the claim’s falsity. However, the rightmost example exposes a critical failure mode: the detector labels a subtly manipulated or synthetic image as REAL with high confidence, despite the claim being false. This erroneous authenticity cue biases the downstream reasoning process, causing the system to underweight semantic and contextual inconsistencies and ultimately misclassify the claim as REAL. This case concretely illustrates how artifact-centric priors can override evidence-based reasoning when pixel-level manipulations are subtle or fall outside the detector’s learned artifact space. This suggests deepfake detector outputs should be treated carefully as an auxiliary hint rather than a deciding factor—and balanced against evidence consistency and contextual checks during claim verification. 

\section{Case Study}
\begin{figure}[!htbp]
\centering
\includegraphics[width=0.9\textwidth]{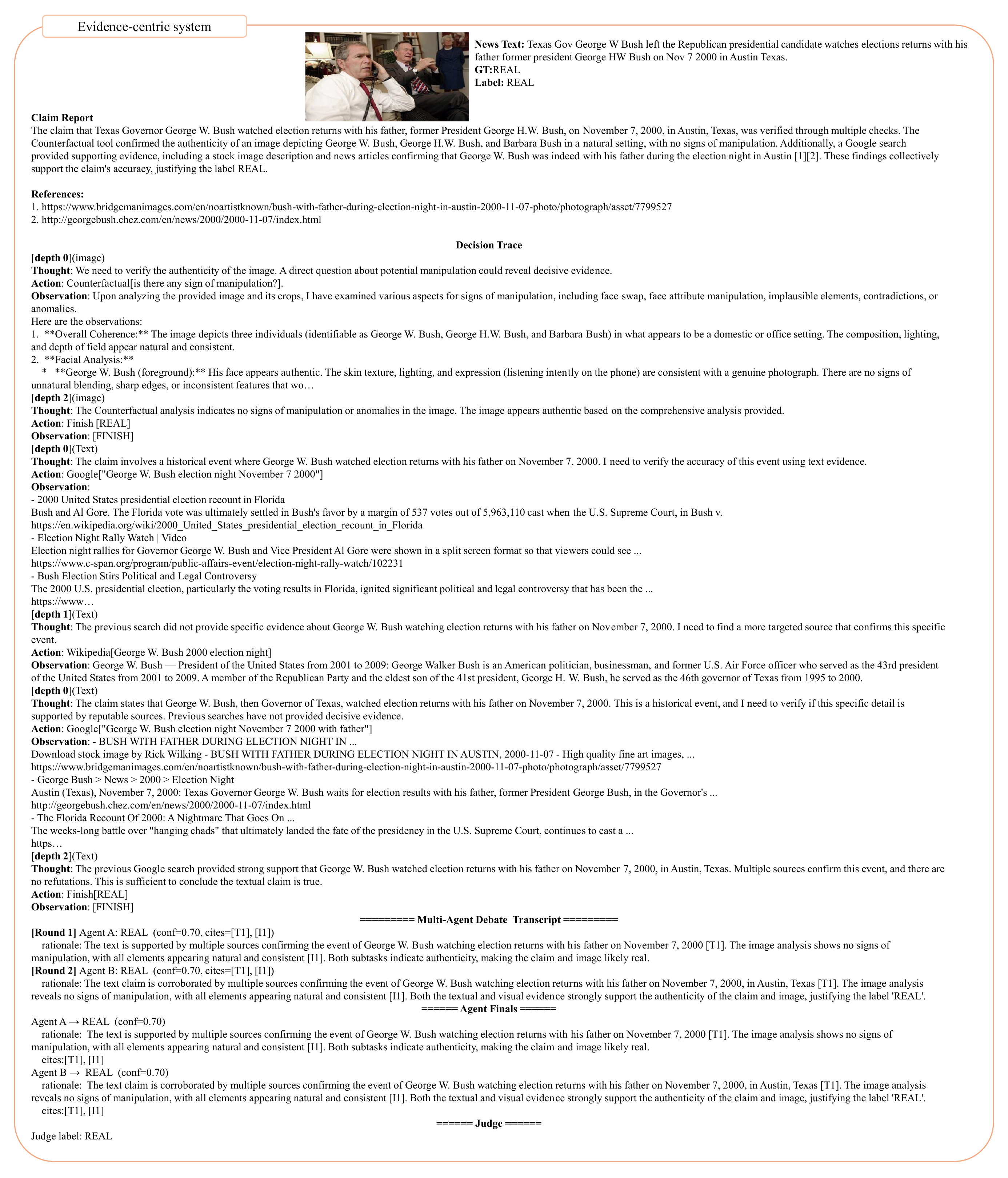}
\caption{The case study of correct REAL classification through evidence-centric fact-checking, combining external source verification with image consistency checks.}
\label{fig:7}
\end{figure}

In the first example shown in Figure~\ref{fig:7}, the fact-checking system correctly classifies a REAL claim describing Texas Governor George W. Bush watching election returns with his father on November 7, 2000. As illustrated in the figure, the image depicts a plausible historical setting with recognizable individuals and a natural indoor context, which is consistent with the claim. Rather than relying on visual appearance alone, the system retrieves multiple independent textual sources, including archival news descriptions and stock image records, which explicitly confirm the event, its date, and location. Image analysis tools further find no indications of manipulation, reinforcing but not determining the assessment. During the Multi-Agent Debate, both verifier agents cite corroborating textual evidence alongside the image consistency checks and identify no temporal or contextual contradictions. The Judge therefore issues a REAL verdict. This example demonstrates how the proposed evidence-centric framework uses the image as contextual support while grounding the final decision primarily in verifiable external evidence and historical records.

\begin{figure}[!htbp]
\centering
\includegraphics[width=0.9\textwidth]{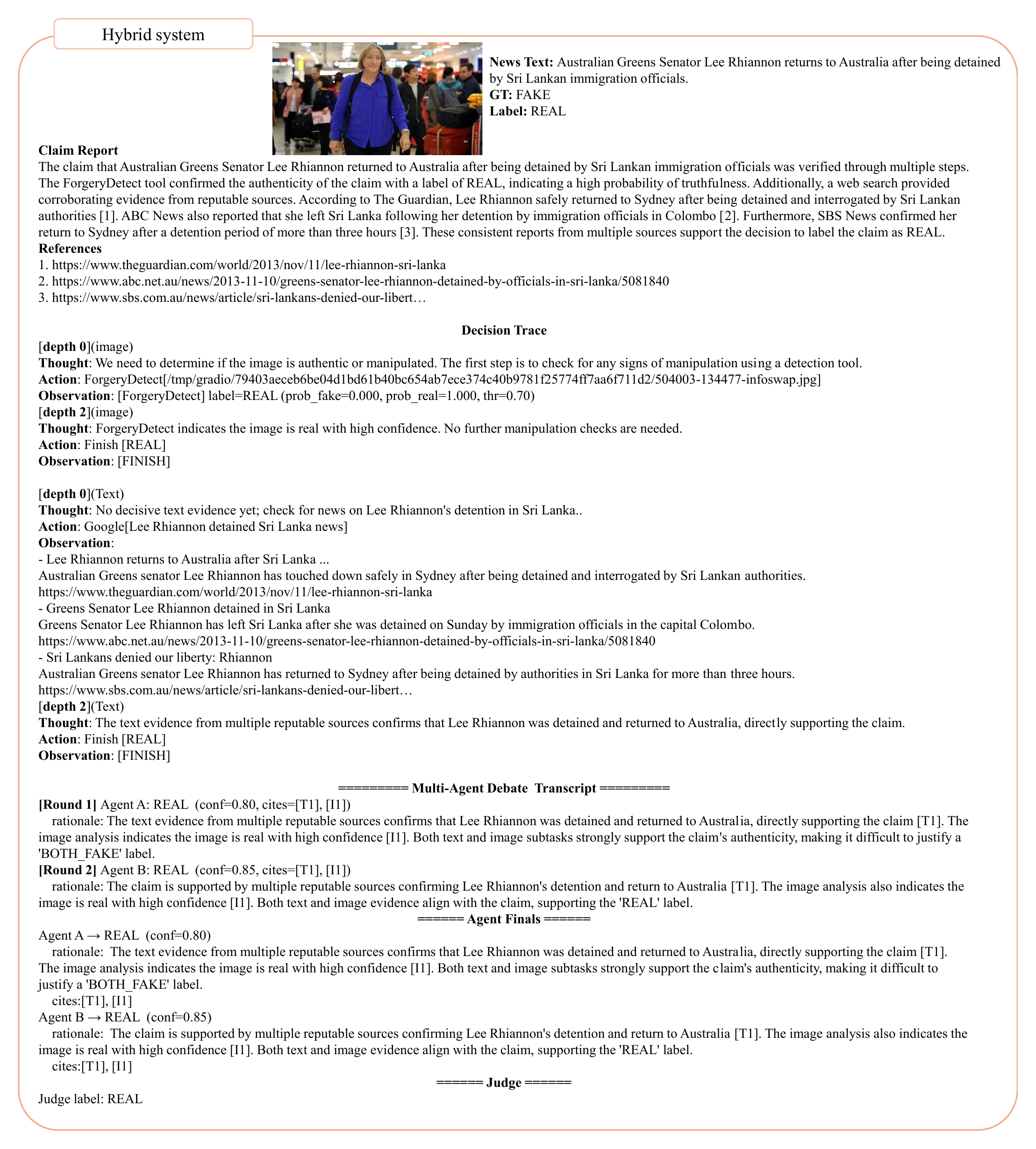}
\caption{Case study of hybrid fact-checking system misclassifies a FAKE claim after the deepfake detector predicts the image as REAL which highlights the need to treat detector signals carefully in the fact-checking system.}
\label{fig:8}
\end{figure}

Figure~\ref{fig:8} illustrates a representative failure case of the hybrid system, where the ground-truth label is FAKE due to image manipulation performed using an InfoSwap-based face-swapping method, but the system ultimately predicts REAL. As shown in the figure, the manipulated image appears visually natural, with consistent lighting, facial structure, and background context, making the swap difficult to detect through surface-level inspection. As InfoSwap produces highly realistic results with minimal pixel-level artifacts, the deepfake detector assigns a confident REAL label to the image. When this detector output is incorporated into the fact-checking pipeline, it acts as an authenticity cue that subtly biases the reasoning process. In the decision trace, the system terminates visual analysis early based on the detector’s high-confidence prediction, while subsequent textual evidence, although factually correct in isolation does not trigger further scrutiny of image authenticity. During the Multi-Agent Debate, both verifier agents reference the detector output alongside corroborating text sources, and the Judge issues a REAL verdict. This case highlights how artifact-centric signals, when faced with realistic face-swapping techniques, may inadvertently reinforce incorrect assumptions about image authenticity, underscoring the importance of treating deepfake detector outputs as auxiliary signals rather than decisive evidence within multimodal claim verification

\section{Conclusion}

This work examined a central question in multimodal misinformation detection: whether image-based deepfake detectors provide any meaningful value for verifying image–text claims, either as standalone predictors or as auxiliary signals within an evidence-centric fact-checking system or whether their artifact-centric predictions instead introduce misleading priors that degrade claim-level reasoning. To answer this question, we evaluated five state-of-the-art image-only detectors alongside our MCTS-guided, tool-augmented fact-checking system with Multi-Agent Debate across MMFakeBench and DGM$^{4}$.
Across both benchmarks, the evidence-centric fact-checking framework consistently outperforms all detector baselines, achieving F1~$\approx$~0.81 on MMFakeBench and  F1~$\approx$~0.55 on DGM$^{4}$, compared with F1 values of only 0.33--0.53 for image-based detectors. Moreover, integrating detector outputs into the fact-checking pipeline systematically \emph{reduces} performance by 0.06--0.08 F1 on MMFakeBench and by roughly 0.04 F1 on DGM$^{4}$ and increases both false negatives and false positives. These findings demonstrate that multimodal misinformation verification is inherently evidence-centric: the truth of an image--text claim is determined primarily by its semantic content and contextual grounding, not by pixel-level artifact signals.
Our findings suggest that although image-based deepfake detectors are commonly deployed as upstream authenticity filters in real-world moderation systems and occasionally incorporated into multimodal fact-checking pipelines, they should not be treated as primary evidential signals for determining the truth of an image–text claim. Effective verification systems must instead prioritize targeted retrieval, provenance, and temporal checks, and deliberative adjudication. Artifact-level cues, when present, should be treated only as optional and non-binding auxiliary signals rather than decisive indicators of claim authenticity.

 \bibliographystyle{elsarticle-num} 
 \bibliography{reference}

\end{document}